\documentclass[letterarticle, 10 pt, conference]{ieeeconf}
\IEEEoverridecommandlockouts
\usepackage{balance}
\usepackage{amssymb}
\usepackage{graphics}
\usepackage{graphicx}
\usepackage{amsmath}
\usepackage{floatrow}
\usepackage{url}
\usepackage{cite}
\usepackage{textcomp}
\usepackage{multirow}
\usepackage{tablefootnote}
\usepackage{siunitx}
\usepackage{tabularray}
\usepackage{xcolor}
\usepackage{array}

\usepackage{hyperref}
\hypersetup{
    colorlinks=true,
    linkcolor=red,
    filecolor=magenta,      
    urlcolor=cyan,
    citecolor=blue
}

\UseTblrLibrary{booktabs}

\title{\LARGE \bf
A ROS 2 Wrapper for Florence-2: Multi-Mode Local Vision-Language Inference for Robotic Systems
}

\author{J. E. Dom\'{i}nguez-Vidal
}

\begin{document}
\maketitle
\thispagestyle{empty}
\pagestyle{empty}

\begin{abstract}
Foundation vision-language models are becoming increasingly relevant to robotics because they can provide richer semantic perception than narrow task-specific pipelines. However, their practical adoption in robot software stacks still depends on reproducible middleware integrations rather than on model quality alone. Florence-2 is especially attractive in this regard because it unifies captioning, optical character recognition, open-vocabulary detection, grounding and related vision-language tasks within a comparatively manageable model size. This article presents a ROS~2 wrapper for Florence-2 that exposes the model through three complementary interaction modes: continuous topic-driven processing, synchronous service calls and asynchronous actions. The wrapper is designed for local execution and supports both native installation and Docker container deployment. It also combines generic JSON outputs with standard ROS~2 message bindings for detection-oriented tasks. A functional validation is reported together with a throughput study on several GPUs, showing that local deployment is feasible with consumer grade hardware. The repository is publicly available here: \url{https://github.com/JEDominguezVidal/florence2_ros2_wrapper}.
\end{abstract}

\begin{keywords}
ROS 2, Foundation Models, Vision-Language Models, Florence-2, Robotic Perception
\end{keywords}

\section{Introduction}\label{sec:introduction}

Recent progress in foundation models has broadened the range of perception and reasoning capabilities that can be brought into robotic systems, replacing solutions based on Deep Learning architectures specifically designed for each individual task~\cite{fiorini2021daily, zafar2023empowering, dominguez2024anticipation, dominguez2025inference}. In particular, vision-language models have made it possible to move beyond fixed-category perception towards more flexible semantic descriptions, open-vocabulary detection and language-conditioned scene understanding~\cite{florence2, florence, clip, groundingdino}. In robotics, this trend is reflected both in embodied multimodal systems such as PaLM-E and RT-2 and in open frameworks such as OpenVLA and VoxPoser, which demonstrate the value of connecting rich perceptual representations with robot behaviour~\cite{palme, rt2, openvla, voxposer}. This need is particularly relevant in collaborative robotics, where perception modules may support not only object-level recognition but also higher-level reasoning about task context, human activity, and their intention~\cite{kim2024understanding, li2025sensorllm, dominguez2025human}.

Despite this progress, the practical adoption of such models in robot software stacks still depends on integration effort. A model that is straightforward to evaluate in Python notebooks is not automatically usable in a ROS~2 graph with camera topics, services, actions, launch files, standard message types and reproducible deployment. This gap between model availability and system usability has been repeatedly observed in ROS and ROS~2 wrapper articles, where the contribution lies less in proposing a new model and more in making an existing capability reusable within robotic software infrastructures~\cite{gymgazebo, gymgazebo2, fogros2, dominguez2024force, nerfbridge}.

Florence-2 is a particularly interesting case. It provides a unified prompt-based interface for a broad set of computer vision and vision-language tasks while remaining substantially easier to deploy than many very large multimodal systems~\cite{florence2}. This makes it an appealing candidate for local robotic perception, especially in settings where internet dependence is undesirable and hardware resources are limited but not negligible. At the same time, unlike speech or segmentation models for which ROS and ROS~2~\cite{wrapyfi, rosllm, rosa, ramirez2023whisper} wrappers are already relatively easy to find, there is still a lack of focused ROS~2 integrations for Florence-2.

This article addresses that gap by presenting a ROS~2 wrapper for Florence-2 intended as a practical software component for robotic systems. The wrapper subscribes to image topics, supports on-demand inference through both services and actions, and can optionally process incoming frames continuously. It is packaged for local execution and Docker-based deployment, and it publishes both general structured outputs and ROS-native detection messages.

The main contributions of this work are as follows:
\begin{enumerate}
    \item an open ROS~2 wrapper for Florence-2 oriented towards local robotic deployment;
    \item a multi-mode interaction design combining continuous processing, synchronous services and asynchronous actions;
    \item a unified interface for several Florence-2 task families within a single ROS~2 node; and
    \item an initial functional and performance validation, including a cross-GPU throughput comparison.
\end{enumerate}

The remainder of the article is organised as follows. Section~\ref{sec:related_work} summarises the most relevant related work. Section~\ref{sec:architecture} describes the wrapper architecture and ROS~2 interfaces. Section~\ref{sec:implementation} outlines the implementation and deployment choices. Section~\ref{sec:evaluation} reports the experimental validation. Section~\ref{sec:limitations} presents limitations and future work, and Section~\ref{sec:conclusion} concludes the article.

\section{Related Work}\label{sec:related_work}

The most relevant literature for this article lies at the intersection of three lines of work. The first is the progression of foundation perception models themselves, including Florence and Florence-2 for unified visual representations~\cite{florence, florence2}, Whisper for robust speech recognition~\cite{whisper}, Segment Anything for promptable image segmentation~\cite{sam}, and open-vocabulary or language-grounded perception models such as CLIP~\cite{clip, mobileclip} and Grounding DINO~\cite{groundingdino}. The second line concerns robotics applications of large multimodal models, including embodied language and vision systems such as PaLM-E, RT-2, OpenVLA and VoxPoser~\cite{palme, rt2, openvla, voxposer}. These works highlight the relevance of rich perceptual back-ends for robot decision making, manipulation and scene understanding.

The third line is the growing ecosystem of ROS and ROS~2 wrappers, bridges and middleware-oriented software components. Earlier examples such as gym-gazebo and gym-gazebo2 framed ROS and ROS~2 integration as a reusable software contribution in its own right~\cite{gymgazebo, gymgazebo2}. More recent work has extended this pattern to cloud robotics platforms, middleware wrappers and embodied AI frameworks, including FogROS2, Wrapyfi, ROS-LLM, ROSA and NerfBridge~\cite{fogros2, wrapyfi, rosllm, rosa, nerfbridge}. In parallel, model-specific wrappers have appeared in repositories and technical reports for capabilities such as Whisper and SAM, for example \texttt{ros2\_whisper}\footnote{\url{https://github.com/ros-ai/ros2_whisper}}, \texttt{ros\_sam}\footnote{\url{https://github.com/robot-learning-freiburg/ros_sam}} and \texttt{ros2\_sam}\footnote{\url{https://github.com/ros-ai/ros2_sam}}. However, to the best of our knowledge, there is not yet a dedicated ROS~2 wrapper article or widely adopted ROS~2 package focused on Florence-2.

The present work therefore occupies a narrow but useful position: it does not propose a new foundation model or a full embodied AI framework, but rather a reusable ROS~2 integration for a compact and capable vision-language model that is well suited to local robotic deployment.

\section{System Design and ROS~2 Architecture}\label{sec:architecture}

\subsection{Overall Node Architecture}
\begin{figure}[t]
    \centering
    \includegraphics[width=0.96\textwidth]{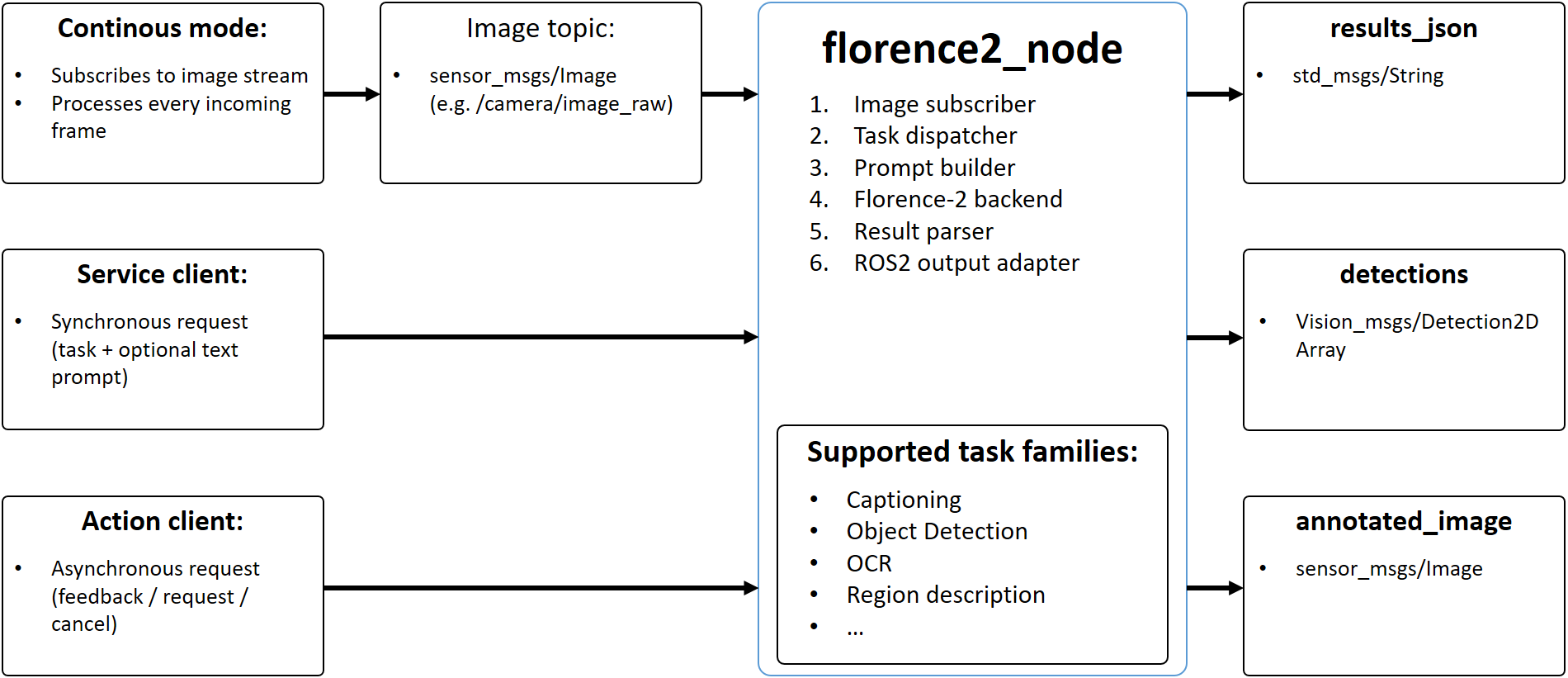}
    \caption{{\bf Architecture of the proposed Florence-2 ROS~2 wrapper.} Diagram showing the main node, its ROS~2 interfaces, and the flow from image acquisition to model inference and result publication.}
    \label{fig:architecture}
\end{figure}

The wrapper is centred on a single ROS~2 inference node that encapsulates model loading, image reception, prompt construction, inference, post-processing and publication of outputs (see Fig.~\ref{fig:architecture}). Internally, the node subscribes to a configurable image topic of type \texttt{sensor\_msgs/Image}, converts incoming frames into a format suitable for Florence-2, runs the model through the Hugging Face \texttt{transformers} interface, and publishes the resulting outputs back into the ROS~2 graph.

This design keeps the runtime path short and makes the wrapper easy to integrate into existing camera-based pipelines. At the same time, it leaves room for future extension through additional pre-processing, batching or task-specific output adapters. The current article focuses on the reusable ROS~2 component rather than on embedding Florence-2 inside a larger autonomous pipeline.

\subsection{Interaction Modes}
A key design choice in the wrapper is the exposure of three complementary interaction modes. First, the node can operate in a \emph{continuous mode}, in which a configured Florence-2 task is executed automatically on every incoming image. This mode is useful when the wrapper is part of an ongoing perception stream and a robot requires a regular semantic interpretation of the camera feed.

Secondly, the wrapper offers a \emph{service mode} for synchronous on-demand inference. This is appropriate when a client node only needs a result occasionally, for example after a waypoint is reached or when a higher-level planner requests a specific perceptual query. Thirdly, the wrapper exposes an \emph{action mode} for asynchronous execution with intermediate feedback. This mode is better aligned with potentially longer inference requests, because it allows clients to monitor progress and integrate Florence-2 into larger task-level execution flows.

From a robotics perspective, this multi-mode design avoids imposing a single interaction pattern on all applications. Continuous operation suits streaming perception, services suit short event-triggered queries, and actions suit longer or better-instrumented requests. The result is a more idiomatic ROS~2 interface than a wrapper limited to a single topic or a single remote procedure style.

\subsection{ROS~2 Interfaces and Message Design}
The wrapper exposes a small but expressive ROS~2 interface surface. The node accepts a configurable image topic, a model selection parameter and an optional continuous task parameter. For on-demand use, it provides the \texttt{ExecuteTask} service and the \texttt{ExecuteTask} action. In both cases the request includes the target Florence-2 task, optional task-specific text input and, when desired, an image payload; otherwise the node can fall back to the most recent subscribed image.

\begin{table*}[t]
\caption{Main ROS~2 interfaces exposed by the wrapper.}
\label{tab:interfaces}
\centering
\begin{tblr}{width=\linewidth,colspec={X[1.8,l]X[0.9,l]X[0.9,l]X[2.4,l]},row{1}={font=\bfseries},rows={valign=m}}
\toprule
Interface & Type & Direction & Purpose \\
\midrule
\texttt{/camera/image\_raw} (configurable) & Topic & Input & Source image stream for continuous or on-demand inference \\
\texttt{\string~/execute\_task} & Service & In/Out & Synchronous task execution with direct response \\
\texttt{\string~/execute\_task\_action} & Action & In/Out & Asynchronous task execution with feedback and result \\
\texttt{\string~/results\_json} & Topic & Output & Generic structured result serialised as JSON \\
\texttt{\string~/detections} & Topic & Output & Standard \texttt{vision\_msgs/Detection2DArray} output for box-based tasks \\
\texttt{\string~/annotated\_image} & Topic & Output & Visualisation of detection-style outputs over the input image \\
\bottomrule
\end{tblr}
\end{table*}

Table~\ref{tab:interfaces} summarises the main interfaces. A noteworthy aspect of the design is the combination of generic and typed outputs. Since Florence-2 supports heterogeneous tasks whose outputs range from plain text to structured detections, a purely typed ROS representation would either be too narrow or would require a large number of bespoke message definitions. The present implementation therefore publishes a generic JSON representation for all tasks, while additionally providing a standard \texttt{vision\_msgs/Detection2DArray} binding and an annotated image for tasks that yield bounding boxes and labels.

This hybrid approach favours broad task coverage without abandoning ROS-native interoperability where it is most useful. In practice, it supports both rapid experimentation and downstream integration with existing detection-oriented nodes and visualisation tools.

\subsection{Task Abstraction}
Florence-2 unifies multiple perception capabilities through prompt-based task tokens~\cite{florence2}. The wrapper preserves this abstraction rather than hard-coding a separate node for each individual capability. As a result, a single ROS~2 component can be used for object detection, captioning, OCR, detailed captioning and related tasks by changing the requested prompt.

This decision is especially relevant for robotics. Instead of maintaining several partially overlapping model servers, developers can expose one model-centric component whose behaviour is selected by the calling node according to context. Such a design is more compact, easier to deploy, and better aligned with the increasingly general nature of foundation perception models.

\section{Implementation and Deployment}\label{sec:implementation}

The software is organised into two packages. The first package contains the custom ROS~2 interfaces, namely the \texttt{ExecuteTask.srv} and \texttt{ExecuteTask.action} definitions. The second package contains the Florence-2 node itself, launch files and example clients for service-based and action-based use. This separation keeps the communication contracts explicit and makes it easier to reuse the interfaces independently of future implementation changes.

The runtime back-end is implemented in Python on top of \texttt{rclpy}, \texttt{torch} and the Hugging Face \texttt{transformers} stack. Model loading selects CPU or GPU execution depending on hardware availability, using reduced precision on CUDA-capable devices where appropriate. Images are converted through \texttt{cv\_bridge}, and object-detection style outputs are converted into \texttt{vision\_msgs} messages when the parsed Florence-2 output contains bounding boxes and labels.

From a deployment standpoint, the wrapper supports two main usage paths. The first is a native local installation inside a Python virtual environment within a standard ROS~2 workspace. The second is based on Docker, including both a lighter configuration and a CUDA-oriented configuration for self-contained GPU deployment. This is an important practical aspect of the contribution: for model wrappers, dependency management and deployment reproducibility are often part of the core value rather than an afterthought~\cite{fogros2, rosllm, rosa}.

The repository also includes example clients that demonstrate both service and action usage. Although simple, these examples are useful because they make the communication contract explicit and reduce the amount of reverse engineering required by new users.

\section{Experimental Validation}\label{sec:evaluation}

\subsection{Experimental Setup}
The purpose of the evaluation is not to benchmark Florence-2 as a vision model against unrelated perception articles, but to validate the wrapper as a ROS~2 software component. All experiments should therefore be interpreted as end-to-end measurements of the proposed integration. The software stack used in this work is based on Ubuntu~24.04, ROS~2 Jazzy, Python~3.12 and the Florence-2 implementation provided through the Hugging Face \texttt{transformers} ecosystem.

Image inputs were provided through standard ROS~2 image topics. For performance measurements, the recommended protocol is to use a repeatable image stream or rosbag replay so that all tested devices process the same input sequence under the same ROS~2 configuration.

\subsection{Functional Validation Across ROS~2 Modes}
The wrapper was functionally validated in its three supported modes. In continuous mode, the node processed incoming images automatically and published generic JSON outputs together with detection-specific outputs when applicable. In service mode, the node returned a direct response for a requested Florence-2 task using either an explicitly provided image or the most recent subscribed image. In action mode, the node additionally emitted execution feedback messages and returned a final result object after inference.

The current implementation accepts action cancellation requests, but the actual interruption granularity is limited by the blocking nature of the underlying generation call. This behaviour should be borne in mind when integrating the action mode into time-critical systems.

\subsection{Performance Observations on Different GPUs}
In addition to functional validation, we are also reporting end-to-end throughput measurements for the wrapper under continuous image processing. The main purpose of this experiment is to show how the practical operating range changes across different hardware profiles.

\begin{table}
\caption{Performance obtained with different Graphic Cards for the Object Detection (OD) task}
\centering
\label{table:performance-graphic-cards}
\begin{tabular}{lcc} 
\hline
\multirow{2}{*}{Graphic Card} & \multicolumn{2}{c}{FPS (min. / \textbf{avg.} / max.)}  \\ \cline{2-3} 
                       & \begin{tabular}[c]{@{}c@{}}Base model\end{tabular} &
                       \begin{tabular}[c]{@{}c@{}}Large model\end{tabular} \\ 
\hline
\begin{tabular}[c]{@{}l@{}}GTX 1060 \\Mobile (80 W)\end{tabular} & 5.50 - \textbf{5.81} - 5.99                                                                  & 2.44 - \textbf{2.50} - 2.56                                                 \\
\begin{tabular}[c]{@{}l@{}}RTX 3060 \\Mobile (80 W)\end{tabular} & 9.23 - \textbf{9.75} - 10.1                                                                 & 4.05 - \textbf{4.21} - 4.29                                                 \\
\begin{tabular}[c]{@{}l@{}}RTX 3080 Ti\\Desktop\end{tabular}     & 25.3 - \textbf{26.6} - 27.5                                                                 & 11.1 - \textbf{11.5} - 11.7                                                \\
\hline
\end{tabular}
\end{table}

Table~\ref{table:performance-graphic-cards} presents these measurements. Different GPUs are used to execute the same oepration (OD) in continuous mode over the same rosbag. Since these cards correspond to different generations, we could not use the same drivers in all of them. Instead, the most recent driver as well as the latest version of CUDA compatible with each driver has been used with each one. As can be seen, an RTX~3060 Mobile would be needed to run the base model at around $10$~FPS for this task.




\section{Limitations and Future Work}\label{sec:limitations}

The current implementation has several limitations. First, not all Florence-2 outputs are mapped to strongly typed ROS~2 messages; JSON remains the universal representation. Secondly, although the action server accepts cancellation requests, interruption is limited by the blocking behaviour of the generation stage. Thirdly, the present validation focuses on integration behaviour and throughput rather than on an exhaustive task-by-task quality assessment. Finally, practical performance remains hardware-dependent, especially for more demanding Florence-2 variants.

Future work will therefore focus on three directions. The first is to extend typed ROS~2 bindings beyond detection-oriented outputs, for example for OCR or region-based tasks. The second is to improve runtime behaviour through optimisation, quantisation or alternative back-ends. The third is to evaluate the wrapper in more complete robotic systems, including navigation, manipulation and human-robot interaction pipelines that can exploit Florence-2 as a local semantic perception component.

\section{Conclusion}\label{sec:conclusion}

This article has presented a ROS~2 wrapper for Florence-2 aimed at practical local deployment in robotic systems. The wrapper exposes the model through continuous, service-based and action-based interaction modes, supports multiple Florence-2 task families through a unified interface, and combines generic structured outputs with ROS-native detection bindings. A small study testing different GPUs throughput is also included. Overall, the contribution is intended as a reusable software bridge that makes a capable vision-language foundation model more accessible to the ROS~2 robotics community.

\bibliographystyle{IEEEtran}
\balance
\bibliography{IEEEabrv,florence2_refs}

\end{document}